\documentclass[sigconf]{acmart}
\usepackage{algorithm}
\usepackage{algorithmic}
\usepackage{multirow}
\usepackage{newfloat}
\usepackage{listings}

\usepackage{array}
\theoremstyle{definition}
\newtheorem{definition}{Definition}

\newcolumntype{L}[1]{>{\raggedright\arraybackslash}p{#1}}
\newcolumntype{C}[1]{>{\centering\arraybackslash}p{#1}}
\newcolumntype{R}[1]{>{\raggedleft\arraybackslash}p{#1}}

\AtBeginDocument{%
  \providecommand\BibTeX{{%
    \normalfont B\kern-0.5em{\scshape i\kern-0.25em b}\kern-0.8em\TeX}}}

\copyrightyear{2024} 
\acmYear{2024} 
\setcopyright{rightsretained} 
\acmConference[KDD '24]{Proceedings of the 30th ACM SIGKDD Conference on Knowledge Discovery and Data Mining}{August 25--29, 2024}{Barcelona, Spain}
\acmBooktitle{Proceedings of the 30th ACM SIGKDD Conference on Knowledge Discovery and Data Mining (KDD '24), August 25--29, 2024, Barcelona, Spain}\acmDOI{10.1145/3637528.3672006}
\acmISBN{979-8-4007-0490-1/24/08}

\begin{document}

\title{Spuriousness-Aware Meta-Learning for Learning Robust Classifiers}
\author{Guangtao Zheng}
\orcid{https://orcid.org/0000-0002-1287-4931}
\affiliation{%
  \institution{University of Virginia}
  \city{Charlottesville}
  \state{VA}
  \country{USA}
  \postcode{22904}
}
\email{gz5hp@virginia.edu}

\author{Wenqian Ye}
\orcid{https://orcid.org/0000-0002-6069-5153}
\affiliation{%
  \institution{University of Virginia}
  \city{Charlottesville}
  \state{VA}
  \country{USA}}
\email{wenqian@virginia.edu}

\author{Aidong Zhang}
\orcid{https://orcid.org/0000-0001-9723-3246}
\affiliation{%
  \institution{University of Virginia}
  \city{Charlottesville}
  \state{VA}
  \country{USA}}
\email{aidong@virginia.edu}







\renewcommand{\shortauthors}{Guangtao Zheng, Wenqian Ye, \& Aidong Zhang}
\newcommand{\ourmethod}{SPUME}
\begin{abstract}
Spurious correlations are brittle associations between certain attributes of inputs and target variables, such as the correlation between an image background and an object class. Deep image classifiers often leverage them for predictions, leading to poor generalization on the data where the correlations do not hold. Mitigating the impact of spurious correlations is crucial towards robust model generalization, but it often requires annotations of the spurious correlations in data -- a strong assumption in practice. In this paper, we propose a novel learning framework based on meta-learning, termed SPUME -- SPUriousness-aware MEta-learning,  to train an image classifier to be robust to spurious correlations. We design the framework to iteratively detect and mitigate the spurious correlations that the classifier excessively relies on for predictions. To achieve this, we first propose to utilize a pre-trained vision-language model to extract text-format attributes from images. These attributes enable us to curate data with various class-attribute correlations, and we formulate a novel metric to measure the degree of these correlations' spuriousness. Then, to mitigate the reliance on spurious correlations, we propose a meta-learning strategy in which the support (training) sets and query (test) sets in tasks are curated with different spurious correlations that have high degrees of spuriousness. By meta-training the classifier on these spuriousness-aware meta-learning tasks, our classifier can learn to be invariant to the spurious correlations.  
We demonstrate that our method is robust to spurious correlations without knowing them a priori and achieves the best on five benchmark datasets with different robustness measures.
\end{abstract}




\maketitle

\section{Introduction}
Spurious correlations are prevalent in real-world datasets. They are brittle associations between certain input attributes and the corresponding target variables. For example, the class \texttt{cow} is correlated with \texttt{grassland} when most training images show a cow on a grassland, but the correlation breaks when a cow is at a beach \cite{geirhos2020shortcut,beery2018recognition}. The grassland feature is spurious as it does not always correlate with the label \texttt{cow} and is not truly predictive for all cow images. Deep image classifiers often use spurious correlations as their prediction shortcuts \cite{geirhos2020shortcut}, such as inferring an image as representing a \texttt{cow} by focusing on the grassland background of the image. Although this shortcut learning strategy can achieve high overall performance when the majority of samples have spurious correlations, it generalizes poorly on samples where spurious correlations do not hold.
Thus, mitigating the reliance on spurious correlations is crucial for obtaining robust image classifiers.

Existing approaches require annotations of spurious correlations or group labels, which separate data into multiple groups with each containing samples of the same class and sharing the same attribute. For example, a group label (cow, grass field) represents all cow images with grass fields as the background. The group labels are used to formulate new optimization objectives \cite{sagawa2019distributionally} or used for model selection and/or model fine-tuning \cite{nam2022spread,nam2020learning,kirichenko2022last,izmailov2022feature}.  However, knowing the group labels in data requires expert knowledge and costly human annotations, which cannot scale to large datasets. Completely removing the requirement for group labels while learning robust classifiers is also a challenging task since we have no knowledge about what spurious correlations we need to mitigate.

In this paper, we propose a novel learning framework to train an image classifier to be robust to spurious correlations without the need of group labels. We design our framework to iteratively detect and mitigate the spurious correlations that the classifier heavily relies on for predictions. To achieve this, we first propose an automatic spurious attribute detection method empowered by a  pre-trained vision-language model (VLM). The VLM enables us to detect text-format attributes which represent many similar pixel-level features and are interpretable to humans.  These attributes together with class labels can formulate various class-attribute correlations which we may find to be spurious in data, and these correlations can cover many potential scenarios where an image classifier fails to generalize because of its reliance on one or multiple of these spurious correlations. 
Therefore, to train a robust classifier against spurious correlations in general without the guidance of group labels,  we focus on mitigating the classifier's reliance on the detected correlations. 

However, it is not efficient to mitigate all of them with equal importance, since among the detected correlations, some are trivial for the classifier as the classifier is robust to them, while some may pose a great risk to the robustness of the classifier. Thus, we propose a novel spuriousness metric to quantify the \textit{spuriousness} of the correlation between a detected attribute and a class label, which measures a classifier's reliance on these class-attribute correlations for predictions, with a larger value indicating a greater reliance on the correlation. With the spuriousness metric, we can identify harmful spurious correlations.

To train a robust classifier, we propose a SPUriousness-aware MEta-learning (termed SPUME) strategy. Unlike the classical settings where only a few spurious correlations are known and needed to be mitigated, our setting has numerous correlations established by the detected attributes and class labels, especially when the dataset that we use has rich features. Using meta-learning, we can distribute the detected spurious correlations with high spuriousness values into multiple meta-learning tasks by carefully curating the data in those tasks. We exploit the support (training) and query (test) sets in a meta-learning task so that samples in the support and query sets have different spurious correlations. Such a task \textit{simulates} a challenging learning scenario where the classifier will perform poorly on the query set when it has a high reliance on the spurious correlations in the support set. By meta-training the classifier on these spuriousness-aware meta-learning tasks,  our classifier can
learn to be invariant to the spurious correlations.

Our \textbf{contributions} are as follows:
\begin{itemize}
    \item We propose an automatic method to detect spurious correlations in data, which exploits the prior knowledge contained in a pre-trained VLM and extracts spurious attributes in interpretable text format.
    \item We tackle the problem of mitigating the reliance on spurious correlations with a novel meta-learning strategy.
    \item  We propose a novel spuriousness metric to guide the construction of meta-learning tasks with the detected spurious attributes.
    \item We demonstrate that a classifier with high average accuracy does not necessarily have high worst-group accuracy which is commonly used for measuring the robustness to spurious correlations.  Our method, termed as \textit{SPUrious-aware MEta-learning (SPUME)}, can train classifiers robust to spurious correlations on five benchmark datasets without knowing the spurious correlations a priori.
\end{itemize}

\section{Related work}
\textbf{Detecting Spurious Attributes.} Spurious attributes spuriously correlate with class labels in data and tend to be exploited for predictions, posing a great risk to the robustness of deep neural classifiers. Detecting spurious attributes typically requires domain knowledge~\cite{clark2019don,nauta2021uncovering} and human annotations \cite{nushi2018towards,zhang2018manifold}. For example, researchers found that object backgrounds \cite{xiao2021noise} and image texture~\cite{geirhos2018imagenettrained} are spurious and can bias the predictions of deep learning models. Recently, model explanation methods \cite{plumb2022finding,abid2022meaningfully} are used to detect spurious attributes.  Neurons in the penultimate layer of a robust model assisted with limited human supervision are also utilized for spurious attribute detection  \cite{singla2021salient,neuhaus2022spurious}. Pre-specifying a set of candidate spurious attributes for spurious attribute detection is also explored  \cite{wu2023discover}. Our method of spurious attribute detection is completely unsupervised. We exploit the prior knowledge in pre-trained VLMs and extract spurious attributes in interpretable text format without any human supervisions.

\noindent\textbf{Mitigating Spurious Correlations.} Spurious correlations tend to bias a model's predictions. There is a growing number of works on mitigating the impact of spurious correlations. Methods that aim to balance data distributions \cite{cui2019class,he2009learning,byrd2019effect} or to perform distributionally robust optimization \cite{sagawa2019distributionally} require knowing group labels which provide information about the spurious correlations in data. Recent works aim to infer group labels to relax this requirement, such as identifying misclassified samples \cite{liu2021just}, clustering hidden representations \cite{pmlr-v162-zhang22z}, invariant learning \cite{creager2021environment}, or training a group label estimator using a small set of data with group labels \cite{nam2022spread}.
\citet{kirichenko2022last} uses group-balanced validation data to retrain the last layer of a model. All these methods still require group labels for the validation data for model selection, which is a strong assumption in practice. A recent work \cite{asgari2022masktune} uses masked data with interpretation techniques to mitigate the impact of spurious correlations without the need of group labels. Our method automatically detects spurious correlations and uses them to construct spuriousness-aware learning tasks and to do model selection. Another line of works is to use data augmentation, such as mixup \cite{zhang2018mixup,han2022umix,wu2023discover} or selective augmentation \cite{yao2022improving}, to mitigate spurious bias in model training. Our method is orthogonal to these approaches as we focus on learning robust classifiers with existing data.

\noindent \textbf{Meta-learning.} Meta-learning \cite{finn2017model,Vinyals2016Matching,Snell2017Prototypical,rusu2018metalearning} is a bi-level learning paradigm and is popular in few-shot learning
\cite{Oreshkin2018TADAM,Sung2018Learning,chen2020new,ye2020fewshot}. It aims to learn from one set of data and to generalize on another set of data. It has been found that the meta-learning can learn high-quality representations \cite{Raghu2020Rapid}, achieving good generalization across different tasks. Utilizing the novel idea of meta-learning, in this paper, we transform the problem of spurious correlation mitigation into a novel meta-learning problem to facilitate learning  feature representations robust to spurious correlations.

\begin{figure*}
    \centering
    \includegraphics[width=0.95\linewidth]{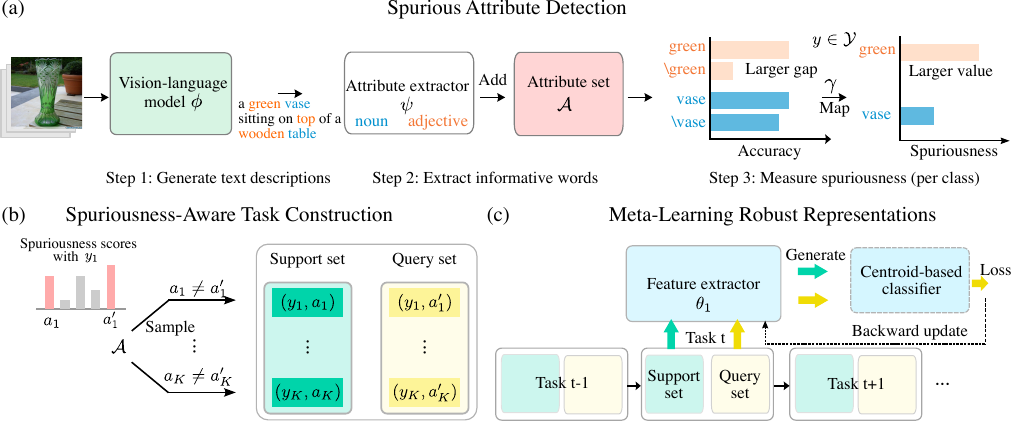}
    \caption{Overview of SPUME. (a) Detect attributes from training data and measure their spuriousness in three steps. ``\textbackslash green" denotes without the attribute ``green". (b) Construct spuriousness-aware meta-learning tasks guided by the spuriousness scores of the detected attributes. (c) Meta-train a robust feature extractor using the constructed tasks.}
    \label{fig:method-overview}
\end{figure*}

\section{Problem Formulation}\label{sec:problem-formulation}
Consider a training dataset $\mathcal{D}_{\text{tr}}=\{(x_n,y_n)\}_{n=1}^N$ with $x_n\in\mathcal{X}$, $y_n\in\mathcal{Y}$, where $\mathcal{X}$ denotes the input space containing all possible inputs, $\mathcal{Y}$ denotes the set of $K$ classes. In real-world scenarios, a sample $x_n$ in $\mathcal{D}_{\text{tr}}$ typically has \textit{spurious attributes} and these attributes have \textit{spurious correlations} with the label $y_n$. We describe the two important concepts below.

\noindent$\bullet$ \textbf{Spurious attributes:} A spurious attribute $a\in\mathcal{A}$ describes some common patterns in the input space $\mathcal{X}$ and spuriously correlates with some label $y\in\mathcal{Y}$, where $\mathcal{A}$ denotes all possible spurious attributes. In other words, $a$ can be in samples of multiple classes or only in some samples of a class, and therefore is not essential to any of the classes. For example, the ``land background" attribute can exist in images of waterbird and landbird classes \cite{sagawa2019distributionally}, and ``land background" is non-essential to either of the classes. 

\noindent$\bullet$  \textbf{Spurious correlations:} A spurious correlation, denoted as $\langle y,a\rangle$, describes the brittle association between the spurious attribute $a$ and the label $y$. The spurious correlation $\langle y,a\rangle$ does not always hold in the sense that $a$ can be associated with multiple $y$'s or $y$ can correlate with other attributes in some samples. Knowing all the spurious correlations in $\mathcal{D}_{\text{tr}}$, we can divide $\mathcal{D}_{\text{tr}}$ into multiple data groups $\mathcal{D}_{\text{tr}}^g$, $g\in\mathcal{G}$, where $g=(y,a)$ denotes the group label for samples with the label $y$ and having the spurious attribute $a$, and $\mathcal{G}=\mathcal{Y}\times \mathcal{A}$ denotes the set of all group labels.

Given a deep neural classifier $f_{\theta}$ with parameters $\theta$, we train it with empirical risk minimization (ERM) on the training set $\mathcal{D}_{\text{tr}}$ and obtain the optimized classifier $f_{\theta^*}$ as follows:
\begin{align}\label{eq:erm-objective}
    \theta^* &=\arg \min_{\theta}\mathbb{E}_{(x,y)\in \mathcal{D}_{\text{tr}}}\ell(f_{\theta}(x),y),
\end{align}
where $\ell(\cdot,\cdot)$ is the cross-entropy loss function. 

The problem occurs when data groups $\{\mathcal{D}_{\text{tr}}^g|g\in\mathcal{G},\mathcal{D}_{\text{tr}}^g\subset\mathcal{D}_{\text{tr}}\}$ in $\mathcal{D}_{\text{tr}}$ are imbalanced in sizes or the inductive bias of the classifier $f_{\theta}$ favors particular data groups. 
For example, a majority group $\mathcal{D}_{\text{tr}}^g$ with the group label $g=(y,a)$ in $\mathcal{D}_{\text{tr}}$, which has significantly more samples than other groups, may bias the optimization in Eq. \eqref{eq:erm-objective} towards favoring the data in $\mathcal{D}_{\text{tr}}^g$  having the spurious correlation $\langle y, a\rangle$, i.e.,
\begin{align}
    \theta^* \approx \arg \min_{\theta}\mathbb{E}_{(x,y)\in \mathcal{D}_{\text{tr}}^g}\ell(f_{\theta}(x),y),\label{eq:erm-group-optimization-approx}
\end{align}
with $|\mathcal{D}_{\text{tr}}^g|\gg |\mathcal{D}_{\text{tr}}^{g'}|$, where $g,g'\in\mathcal{G}$ and $g\neq g'$, and $|\cdot|$ denotes the size of a set.
As a result, the classifier $f_{\theta^*}$, instead of utilizing the core features in samples to predict $y$, may superficially learn the mapping from $a$ to $y$,  which is non-robust when the correlation between $a$ and $y$ breaks. More specifically, since $a$ is a spurious attribute, there may exist $\langle y',a\rangle$ in samples from class $y'$ with $y\neq y'$. Then, it is very likely that $f_{\theta^*}$ will wrongly predict these samples as $y$ instead of $y'$. For example, when $f_{\theta^*}$ learns to use water backgrounds ($a$) to predict waterbirds ($y$), it fails to recognize landbirds ($y'$) with water backgrounds. Similarly, when the inductive bias in $f_{\theta^*}$ favors certain spurious correlations, the classifier will encounter the same generalization problem.

Spurious correlations pose a great challenge to the robustness of machine learning models. To address this, typically, all or partial group labels of the training data is  required for various purposes, such as formulating the group robustness objective \cite{sagawa2019distributionally}, reweighting the training data, or selecting models \cite{liu2021just}. However, acquiring group labels for a dataset typically involves human-guided annotations, which is costly and not scalable, especially when the dataset is large. In the following,  without the need of group labels, we propose a novel spuriousness-aware meta-learning framework to train a classifier to be robust to spurious correlations.

\section{Spuriousness-Aware Meta-Learning}
We give the overview of our framework in Fig. \ref{fig:method-overview}, where we first detect spurious attributes with a pre-trained VLM (Fig. \ref{fig:method-overview}(a) and Section \ref{sec:spurious-detection}). To effectively use the detected spurious attributes for spurious correlation mitigation, we propose a novel meta-learning strategy and provide details on how to construct spuriousness-aware meta-training tasks (Fig. \ref{fig:method-overview}(b) and Section \ref{sec:task-construction}) and meta-learn robust representations (Fig. \ref{fig:method-overview}(c) and Section \ref{sec:meta-learning}).

\subsection{Automatic Spurious Attribute Detection}\label{sec:spurious-detection}
To automatically detect spurious attributes in a target dataset without human-guided annotations, we propose to exploit the prior knowledge in a pre-trained VLM. Our method detects spurious attributes in \textit{text format} and consists of the following three steps.

\paragraph{Step 1: Generate Text Descriptions.} We generate a text description for each image using a pre-trained VLM $\phi$, which is capable of generating text descriptions of images at scale. Moreover, since the model is trained on massive data and is not specifically fine-tuned on the target dataset, it can discover general objects and patterns. For example, in Fig. \ref{fig:method-overview}(a), besides the class object \texttt{vase}, the VLM also detects the vase's color \texttt{green} and a background object \texttt{table} with its material \texttt{wooden}.  
 
\paragraph{Step 2: Extract Informative Words as Attributes.} We extract informative words from the text descriptions of images as attributes. We select nouns, which describe objects, and adjectives, which describe certain properties of objects, as the informative words.  For example, we extract \texttt{green}, \texttt{vase}, \texttt{top}, \texttt{wooden}, and \texttt{table} from the description in Fig. \ref{fig:method-overview}(a). We instantiate the attribute extractor $\psi$ with an automatic procedure  (Section \ref{sec:experimental-setup}) to extract these informative words from the text descriptions obtained in the first step. Then, these extracted words are added to the attribute set $\mathcal{A}$ as the \textit{possible} spurious attributes. 

\noindent\textbf{Remark.} VLMs can detect general objects and patterns. However, due to the inductive bias learned during pre-training, VLMs may generate text descriptions for some images that are not aligned with human understandings, such as describing a red-and-green background as a ``Christmas tree". Although ``Christmas tree" is not self-explanatory in this case, it is still a valid and useful attribute, representing samples having similar red-and-green backgrounds. This also highlights the benefit of using VLMs: they can detect patterns that are not easily perceived by humans. A limitation of such a VLM-based detection approach is that VLMs may struggle on describing images from domain-specific tasks where, for example, slight changes in orientation of objects or variations in geographies are important for robust predictions. Nevertheless, our proposed spurious attribute detection approach is not restricted to a specific VLM, and it can be improved if more capable VLMs are available.

\paragraph{Step 3: Measure Spuriousness.}
To know whether a detected attribute $a\in\mathcal{A}$ is spurious, we need to consider it in the correlation with a class label $y$, since among all the correlations between the attributes in $\mathcal{A}$ and class labels, some of them may be vacuous --- they do not exist in the training data (e.g., $a$ only exists in images of the class $y'$ with $y'\neq y$), and some of them are not spurious (e.g., the attribute $a$ is detected exclusively in all the images of the class $y$). Moreover, we are interested in identifying spurious correlations that are likely to be exploited by a classifier for predictions as these correlations directly affect the robustness of the classifier.

To unify the above cases, we propose a metric to quantify the likelihood of the correlation $\langle y,a\rangle$  being spurious \textit{and} used by a classifier, i.e., \textit{spuriousness} of the correlation. The metric $\gamma$ considers $y$, $a$, the training data $\mathcal{D}_{\text{tr}}$, and the classifier $f_{\theta}$, and maps them to a finite value, which we call \textit{spuriousness score}. We defines $\gamma$ as follows.

\begin{definition}[Spuriousness Metric]
Given a class label $y\in\mathcal{Y}$, an attribute $a\in\mathcal{A}$, and a classifier $f_{\theta}$ trained on $\mathcal{D}$ with $\theta\in\Theta$, the spuriousness metric for $\langle y,a \rangle$ is a mapping $\gamma:\mathcal{Y}\times\mathcal{A}\times \mathcal{D}\times\Theta\rightarrow [\alpha,\beta]$, where $\mathcal{D}$ denotes a set of sample-label pairs, $\Theta$ denotes the set of all possible $\theta$, and $[\alpha,\beta]$ denotes the output value range of $\gamma$, with $\alpha$ being the lowest and  $\beta$ being the highest.  When the data group size $|\mathcal{D}^{(y,a)}|=0$ or $|\mathcal{D}^{(y,\hat{a})}|=0$, where $\hat{a}$ denotes all attributes in $\mathcal{A}$ other than $a$, the mapping $\gamma$ outputs $\alpha$.
\end{definition}

Given the training set $\mathcal{D}_{tr}$, $|\mathcal{D}_{tr}^{(y,a)}|=0$ and $|\mathcal{D}_{tr}^{(y,\hat{a})}|=0$ correspond to that $\langle y,a \rangle$ does not exist in $\mathcal{D}_{tr}$ and that $\langle y,a \rangle$ exists exclusively in samples of class $y$, respectively. For both cases, the spuriousness of $\langle y,a \rangle$ should be the smallest.

Then, we specifically design $\gamma$ based on the performance of the classifier $f_{\theta}$. The motivation is that the classifier $f_{\theta}$ will generalize poorly on samples of the class $y$ without the attribute $a$ if $f_{\theta}$ excessively relies on $a$ for predicting the label $y$. Therefore, as demonstrated in Fig. \ref{fig:method-overview}(a), the spuriousness will be higher if $f_{\theta}$ has a larger performance discrepancy on images with and without  $a$  and be lower when the performance discrepancy is smaller. We formally define our spuriousness metric for $\langle y,a\rangle$ as follows,

\begin{equation}\label{eq:spuriousness-score}
    \gamma(y,a;\mathcal{D}_{tr}, f_{\theta})= \tanh\Big({\text{abs}\big(\log\frac{J(\mathcal{D}_{tr}^{( y,a)};f_{\theta})}{J(\mathcal{D}_{tr}^{( y,\hat{a})};f_{\theta})}\big)}\Big),
\end{equation}
with $\gamma(y,a;\mathcal{D}_{tr}, f_{\theta})=0$ when $\mathcal{D}_{tr}^{(y,\hat{a})}=\emptyset$ or $\mathcal{D}_{tr}^{(y,a )}=\emptyset$, where $\mathcal{D}_{tr}^{(y,a )}\subset\mathcal{D}_{tr}$ denotes the subset of all training data from the class $c$ \textit{with} the attribute $a$, $\mathcal{D}_{tr}^{(y,\hat{a})}\subset\mathcal{D}_{tr}$ denotes the subset of all training data from the class $c$ \textit{without} the attribute $a$, $J(\cdot;f_{\theta})$ denotes the classification accuracy of $f_{\theta}$ on a given set of samples, and abs($\cdot$) denotes taking the absolute value. The division in Eq. \eqref{eq:spuriousness-score} aims to produce larger values than the simple difference between the two accuracies, making different correlations more distinctive. Moreover, using $\log(\cdot)$ avoids encountering extreme values from the division, and $\tanh(\text{abs}(\cdot))$ bounds the score in the range from 0 to 1.  Other designs of $\gamma$ are possible, and we have shown in our experiments that our method proposed in the following is robust to different choices of spuriousness metrics.



\paragraph{Discussion.} 
 With the detected attributes and our spuriousness metric, we can identify spurious correlations that are likely to be used for predictions by a classifier and thus pose a potential risk to the robustness of the classifier.
To improve the robustness to spurious correlations, we need to mitigate the classifier's reliance on those spurious correlations. Since there are multiple spurious correlations, mitigating all of them at once is a challenging task.
To address this, we formulate the problem in a novel \emph{meta-learning}~\cite{finn2017model,Vinyals2016Matching,Snell2017Prototypical, chen2020new} setting, where we construct meta-learning tasks with each task containing some potentially harmful spurious correlations. Now, our goal is to learn a good classifier that performs well across all these tasks with various spurious correlations.

In the following, we first introduce how to construct meta-learning tasks with the identified spurious correlations. Then, we give the details of using the constructed tasks for meta-learning.

\subsection{Spuriousness-Aware Task Construction} \label{sec:task-construction}

 To mitigate spurious correlations via meta-learning, we first create meta-learning tasks which will be used in meta-training. A meta-learning task typically consists of  a support (training) set $\mathcal{S}$ providing training samples for learning novel concepts and a query (test) set $\mathcal{Q}$ containing test samples for the evaluation of the learning outcome.  We use the two sets to \textit{simulate} spurious correlations in meta-learning tasks so that these spurious correlations can be effectively mitigated via meta-learning.

As illustrated in Fig. \ref{fig:method-overview}(b), for \textit{each class} $y_k$ with $k=1,\ldots,K$, we first sample two attributes $a_k$ and $a_k'$ from $\mathcal{A}$ based on their spuriousness scores, where $a_k\neq a_k'$. Specifically, we normalize the scores as probabilities, and an attribute with a higher spuriousness score will be more likely to be selected than another attribute with a lower spuriousness score. In this way, we target the spurious correlations that pose a high risk to the robustness of the classifier.

Then, the two sampled attributes formulate two spurious correlations with $y_k$, i.e., $\langle y_k,a_k\rangle$ and $\langle y_k,a_k'\rangle$, based on which, we get two data groups, $\mathcal{D}_{\text{tr}}^{(y_k,a_k)}$ and $\mathcal{D}_{\text{tr}}^{(y_k,a_k')}$, from the training set $\mathcal{D}_{\text{tr}}$. These two groups of data together represent a shift in the correlation between the two spurious attributes and the class label. If the classifier learns to rely on the spurious correlation in one group of data for predictions, then it will fail on the other group of data with a different spurious correlation. Thus, crafting such a shift facilitates learning a robust classifier.

Next, for efficient training, we randomly sample $N_S$ data points per class from the two data groups to construct the \textit{non-overlapping} support set $\mathcal{S}_k$ and the query set $\mathcal{Q}_k$, i.e.,
\begin{equation}\label{eq:support-set}
    \mathcal{S}_k=\bigcup_{i=1}^{N_S}\big\{(x_i,y_k)|(x_i,y_k)\in\tilde{\mathcal{D}}_{\text{tr}}^{(y_k,a_k)}\big\},
\end{equation}
and
\begin{align}\label{eq:query-set}
    \mathcal{Q}_k=\bigcup_{i=1}^{N_S}\big\{(x_i,y_k)|(x_i,y_k)\in\tilde{\mathcal{D}}_{\text{tr}}^{(y_k,a_k')}\big\},
\end{align}
where $\tilde{\mathcal{D}}_{\text{tr}}^{(y_k,a_k)}=\mathcal{D}_{\text{tr}}^{(y_k,a_k)}-\mathcal{D}_{\text{tr}}^{(y_k,a_k')}$ and $\tilde{\mathcal{D}}_{\text{tr}}^{(y_k,a_k')}=\mathcal{D}_{\text{tr}}^{(y_k,a_k')}-\mathcal{D}_{\text{tr}}^{(y_k,a_k)}$ are sets of elements unique to $\mathcal{D}_{\text{tr}}^{(y_k,a_k)}$ and $\mathcal{D}_{\text{tr}}^{(y_k,a_k')}$, respectively. Taking the above set difference ensures that the two spurious correlations won't appear in the same set since some samples may have both the attributes $a_k$ and $a_k'$. 

After constructing the two sets for \textit{each class}, we obtain the constructed task $\mathcal{T}=\{\mathcal{S},\mathcal{Q}\}$ with $\mathcal{S}=\cup_{k=1}^K\mathcal{S}_k$ and $\mathcal{Q}=\cup_{k=1}^K\mathcal{Q}_k$. If $K$ is large, we can randomly select a subset of $K$ classes to construct $\mathcal{T}$.
The constructed task $\mathcal{T}$ demonstrates to the classifier that the spurious correlations in $\mathcal{T}$ are highly risky for it, and that the classifier should be invariant to them in order to perform well on this task. Importantly, the construction of meta-learning tasks also ensures that biases in VLMs won't be passed down to the classifier as the construction process effectively decorrelates biased attributes from VLMs with prediction targets.

\begin{algorithm}[t]
        \caption{SPUME}
        \begin{flushleft}
        \textbf{Input:} A training dataset $\mathcal{D}_{tr}$, a feature extractor $h_{\theta_1}$, a pre-trained VLM $\phi$, an attribute extractor $\psi$, a spuriousness metric $\gamma$, the number of tasks per epoch $N_T$, the number of classes $K$, and the number of training epochs $E$.\\ 
        \textbf{Output}: Learned weights $\theta_1^*$
        \end{flushleft}
		\begin{algorithmic}[1]
        \STATE{Build the attribute set $\mathcal{A}=\cup_{(x,y)\in\mathcal{D}_{tr}}\psi(\phi(x))$}
        \FOR{$e=1,\ldots,E$}
        \STATE Generate class centroids with Eq. \eqref{eq:class-centroids} using $\mathcal{D}_{tr}$
		\STATE Generate spuriousness scores using Eq. \eqref{eq:spuriousness-score}
  \STATE Set $T(\mathcal{D}_{\text{tr}},\mathcal{A},\gamma,\theta_1)$ as an empty set
          \FOR{$t=1,\ldots,N_T$}
            \STATE Sample $K$ pairs of attributes from $\mathcal{A}$ for each class
            \STATE Construct a spuriousness-aware meta-learning task $\mathcal{T}$ using Eq. \eqref{eq:support-set} and \eqref{eq:query-set}
            \STATE Add $\mathcal{T}$ to $T(\mathcal{D}_{\text{tr}},\mathcal{A},\gamma,\theta_1)$
          \ENDFOR
      \STATE Optimize $\theta_1$ using Eq. \eqref{eq:objective}
        \ENDFOR
        \RETURN $\theta_1^*$ 
		\end{algorithmic}\label{alg:1}
\end{algorithm}

\subsection{Meta-Learning Robust Representations} \label{sec:meta-learning}
To train a robust classifier using the constructed tasks, we modify $f_{\theta}$ so that it fits in with the meta-learning paradigm. Specifically, we discard the last linear classification layer of $f_{\theta}$ and keep its feature extractor $h_{\theta_1}:\mathcal{X}\rightarrow \mathbb{R}^D$, where $\theta_1\subset\theta$ and $D$ is the number of dimensions in the feature extractor's outputs. Thus, learning a robust classifier is equivalent to learning robust representations.

As illustrated in Fig. \ref{fig:method-overview}(c), for the $t$'th task, we use the representations of the samples in the support set $\mathcal{S}$ provided by $h_{\theta_1}$ to generate (learn) a centroid-based classifier with $K$ class-centroids $\mathcal{W}=\{\mathbf{w}_1,\ldots,\mathbf{w}_K\}$ calculated as follows
\begin{align}\label{eq:class-centroids}
    \mathbf{w}_k=\frac{1}{N_S}\sum_{n=1}^{N_S}h_{\theta_1}(x_n), (x_n,y_k)\in\mathcal{S}.
\end{align}
Next, we evaluate whether the classifier depends on the spurious correlations in $\mathcal{S}$ by testing it on the query set $\mathcal{Q}$ where the spurious correlations in $\mathcal{S}$ do not hold. The output probability of the classifier on $y_k$ is calculated as follows
\begin{align}\label{eq:centroid-classifier}
    p(y_k|x_n, \theta_1,\mathcal{S})=\frac{\exp(\tau{d(\mathbf{w}_k,h_{\theta_1}(x_n)))}}{\sum_{k'=1}^K\exp(\tau d(\mathbf{w}_{k'},h_{\theta_1}(x_n)))},
\end{align}
where $d(\cdot,\cdot)$ denotes the cosine similarity between two embedding vectors, and $\tau$ denotes a scaling hyperparameter. Then, the task loss $\ell_{\mathcal{T}}$ on $\mathcal{T}=\{\mathcal{S},\mathcal{Q}\}$ is as follows
\begin{equation}\label{eq:task-loss}
    \ell_{\mathcal{T}}(\theta_1) = \underset{(x_n,y_n)\in\mathcal{Q}}{\mathbb{E}} -\log p(y_n|x_n,\theta_1,\mathcal{S}).
\end{equation}
A high loss indicates that the classifier, and in turn the feature extractor $h_{\theta_1}$, rely on the spurious correlations in the support set and cannot generalize well on the query set. 

\noindent\textbf{Learning Objective.}  We minimize the loss in \eqref{eq:task-loss} over tasks constructed with various spurious correlations to find a feature extractor $h_{\theta_1^*}$ that is robust to multiple spurious correlations, i.e.,
\begin{align}\label{eq:objective}
    \theta_1^*=\arg\min_{\theta_1}\mathbb{E}_{\mathcal{T}\in T(\mathcal{D}_{\text{tr}},\mathcal{A},\gamma,\theta_1)}\ell_{\mathcal{T}}(\theta_1),
\end{align}
where $T(\mathcal{D}_{\text{tr}},\mathcal{A},\gamma,\theta_1)$ denotes all possible meta-learning tasks constructed from $\mathcal{D}_{\text{tr}}$ based on the detected attributes $\mathcal{A}$, the spuriousness metric $\gamma$, and the feature extractor $\theta_1$.

To solve \eqref{eq:objective}, we adopt an iterative optimization procedure. We first fix $\theta_1$ and construct a set of meta-training tasks based on $\mathcal{A}$, $\theta_1$, and $\gamma$. Then, we update $\theta_1$ using the constructed tasks. The above steps are iterated until some stop criterion is met.  We name our method as \textit{SPUriousness-aware MEta-Learning (\ourmethod)} and give the training details in Algorithm \ref{alg:1}. 

\noindent\textbf{Complexity Analysis.} VLMs do not incur training cost because they are only used for data preparation.  Extracting attributes (Line 1, Algorithm \ref{alg:1}) is a onetime offline process, and empirically, its time cost scales linearly with the dataset size. Spuriousness measurement (Line 4, Algorithm \ref{alg:1}) is performed periodically during training, and its time complexity grows linearly with the amount of data it uses.  The total training cost is $O(E (C_{m}+C_{s}))$, where $E$ is the number of training epochs, $C_{m}$ and $C_{s}$ are the time cost of meta-learning a classifier and obtaining spuriousness scores per epoch, respectively, with $C_{m}\gg C_{s}$, since the latter only requires forward passes through the classifier. Moreover, using a metric-based meta-learning technique (Eq. \eqref{eq:class-centroids}) leads to $C_{m}$ being comparable to training a standard classifier. Therefore, our method does not incur significant training cost compared with the ERM training.

\noindent\textbf{Model Selection.} We divide the validation data $\mathcal{D}_{\text{val}}$ into groups based on the detected attributes $\mathcal{A}$ and calculate the average accuracy over these groups as follows,
\begin{equation}\label{eq:pseudo-unbiased accuracy}
   Acc_{pu}= \frac{1}{|\mathcal{A}|\cdot|\mathcal{Y}|}\sum_{a\in\mathcal{A}}\sum_{y\in\mathcal{Y}}J\big(\mathcal{D}_{\text{val}}^{(y,a)};h_{\theta_1}\big).
\end{equation}
We call this metric \textit{pseudo-unbiased accuracy}, which fairly measures the performance of the classifier on various groups inferred with the detected attributes in $\mathcal{A}$.

\noindent\textbf{Inference.}
We first create a centroid-based classifier using Eq. \eqref{eq:class-centroids} with all the data in $\mathcal{D}_{\text{tr}}$. Then, given a test sample $x$, the prediction is $\hat{y}=\arg\max_{y\in\mathcal{Y}}p(y|x,\theta_1,\mathcal{D}_{\text{tr}})$.

\section{Experiment}\label{sec:experiment}


\subsection{Datasets}
We tested our method on five image classification datasets with various types of spurious correlations, which are introduced below. Detailed dataset statistics are give in Table \ref{tab:dataset-statistics} in Appendix.

\noindent\textbf{Waterbirds} \cite{sagawa2019distributionally} contains waterbird and landbird classes. It is a synthetic dataset generated by combining images of the two kinds of birds from the CUB dataset \cite{WelinderEtal2010} with water and land backgrounds  from the Places dataset \cite{zhou2017places}, producing  (landbird, land), (landbird, water), (waterbird, land), and (waterbird, water) groups.  

\noindent\textbf{CelebA} \cite{liu2015deep} is a large-scale image dataset of celebrity faces. It contains images showing two hair colors, non-blond and blond, which are spuriously correlated with gender. There are four groups in the CelebA dataset: (non-blond, female), (non-blond, male), (blond, female), and (blond, male). 

\noindent \textbf{ImageNet-9} \cite{ilyas2019adversarial} is a subset of ImageNet \cite{imagenet} and contains nine super-classes. It is known to have correlations between object classes and image textures.   We followed the setting in \cite{kim2022learning} and \cite{bahng2020learning} to prepare  training and validation data.

\noindent \textbf{ImageNet-A} \cite{hendrycks2021natural} is a dataset of real-world images, adversarially curated to test the limits of classifiers such as ResNet-50. While these images are from standard ImageNet classes \cite{imagenet}, they are often misclassified in multiple models. We used this dataset to test the robustness of a classifier after training it on ImageNet-9.

\noindent \textbf{NICO} \cite{he2021towards} is designed for out-of-distribution image classification, simulating real-world scenarios where testing distributions differ from training ones. It labels images with both main concepts (e.g., cat) and contexts (e.g., at home). We used the Animal super-class in NICO and followed the setting in \cite{bai2021decaug,pmlrv202tiwari23a} for data preparation.

\subsection{Experimental Setup}
\label{sec:experimental-setup}
\paragraph{Spurious Attribute Detection.} We used two pre-trained VLMs, ViT-GPT2 \cite{nlp_connect_2022} and BLIP  \cite{li2022blip} to generate text descriptions for images. ViT-GPT2 has an encoder-decoder structure with a vision transformer \cite{dosovitskiyimage} as the encoder and the language model GPT-2 \cite{radford2019language} as the decoder. BLIP has a multimodal mixture of encoder-decoder architecture. 
After generating text descriptions, we used Spacy (\url{https://spacy.io/}) to extract nouns and adjectives from the descriptions automatically. We additionally filtered out words with frequencies less than 10 to remove potential annotation noise and to ensure that we have enough samples to construct a meta-learning task with selected spurious attributes. We give the statistics of the detected spurious attributes in the four datasets (ImageNet-A is not included as it is only used for testing) in Table \ref{tab:statistics-detected-spurious-attributes}. BLIP detects more attributes than ViT-GPT2 overall but less attributes for each image. Based on the two VLMs, our method has two variations, namely \textbf{SPUME-BLIP} and \textbf{SPUME-ViT-GPT2}. In the following experiments, we report the results of both methods.

\begin{table}[t]
\begin{tabular}{ccccc}
\toprule
\multirow{3}{*}{Dataset}    & \multicolumn{2}{c}{\begin{tabular}[c]{@{}c@{}}Number of \\ detected attributes\end{tabular}} & \multicolumn{2}{c}{\begin{tabular}[c]{@{}c@{}}Average number of \\ attributes per image\end{tabular}}\\ \cmidrule{2-5}
& BLIP& ViT-GPT2 & BLIP& ViT-GPT2\\ \midrule
Waterbirds &160 & 144                                                                      &3.301 &4.314                                                                             \\
CelebA     &683 &345                                                                      & 3.913 &4.291                                                                             \\
NICO       & 239&199                                                                      & 3.104&3.995                                                                             \\
ImageNet-9 &540 &442                                                                      &3.276 &4.311                                                                             \\ \bottomrule
\end{tabular}%
\caption{Statistics of the attributes detected from the Waterbirds, CelebA, NICO, and ImageNet-9 datasets.}\label{tab:statistics-detected-spurious-attributes}
\end{table}

\paragraph{Training Settings.} We set $N_S=10$ for sampling each class of images for both the support and query sets of a task.
Following existing settings \cite{sagawa2019distributionally,kim2022learning,pmlrv202tiwari23a}, we used ResNet-50 as the feature extractor for the experiments on the Waterbirds and CelebA datasets, and used ResNet-18 on the ImageNet-9 and NICO datasets. All models were initialized with weights pre-trained on ImageNet. We used a stochastic gradient descent (SDG) optimizer with a momentum of 0.9 and  a weight decay of $10^{-4}$ during meta-training. We trained a model for 100 epochs and used the cosine annealing scheduler to control the decay of learning rate. 
Without any group labels, our method used the pseudo-unbiased accuracy on the validation set defined in Eq. \eqref{eq:pseudo-unbiased accuracy} for model selection, while other methods used the average validation accuracy.  We repeated each experiment three times and calculated the averaged results with standard deviations. We provide additional training details in Appendix. All experiments were conducted on NVIDIA A100 GPUs. We provide an open-source implementation of our method SPUME at \url{https://github.com/gtzheng/SPUME}.

\paragraph{Baselines.} We compare our methods with state-of-the-art methods on mitigating spurious correlations and provide descriptions of the baseline methods in Appendix. For fair comparison, the same feature extractor was used for methods compared on each dataset. Group labels were not used for model training and selection for all the compared methods. Note that we did not include VLMs as baselines, as they were exclusively used for extracting attributes from training data in our method. Moreover, directly using VLMs requires a completely different design, e.g., designing proper input prompts for classification.

\paragraph{Evaluation Metrics.} To evaluate the robustness to spurious correlations on the Waterbirds and CelebA datasets, which provide group labels, we adopted the widely accepted robustness metric, \textbf{worst-group accuracy}, that gives the lower-bound performance of a classifier on the test set with various dataset biases. We also calculated the \textbf{accuracy gap} between the standard average accuracy and the worst-group accuracy 
as a measure of a classifier's reliance on spurious correlations. A high worst-group accuracy with a low accuracy gap indicates that the classifier is robust to spurious correlations and can fairly predict samples from different groups. We adopted \textbf{average accuracy} for the evaluations on the NICO, ImageNet-9, and ImageNet-A datasets as the these datasets are specifically constructed to evaluate the robustness to distributional shifts.

\subsection{Visualization of a Spuriousness-Aware Task}
\begin{figure}[t]
    \centering
    \includegraphics[width=0.95\linewidth]{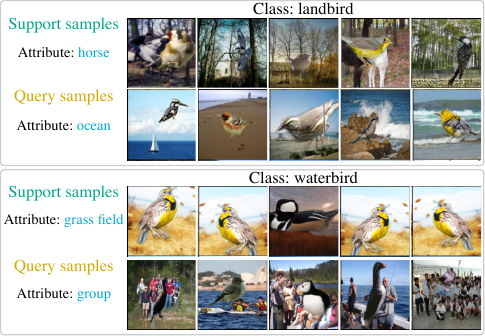}
    \caption{A meta-learning task with $N_S=5$ constructed from the Waterbirds dataset. Images in the support set differ significantly from images in the query set in terms of their backgrounds.}
    \label{fig:task-example}
\end{figure}

We show a spuriousness-aware meta-learning task constructed from the Waterbirds dataset with $N_S=5$ in Fig. \ref{fig:task-example}. For images in the same class, their backgrounds differ significantly in the support and query sets. Specifically, the landbird images selected based on the attribute ``horse" in the support set have land backgrounds, while the same-class images selected based on the attribute ``ocean" in the query set mainly have water backgrounds. Similarly, the query images of waterbird selected based on the attribute ``group" have backgrounds filled with a group of people, while the corresponding support images selected based on the attribute ``grass field" have grass backgrounds without irrelevant objects.

The constructed task creates a challenging learning scenario for classifiers that rely on spurious correlations for predictions. For example, a classifier that learns to use the land backgrounds to predict landbird from the support set will fail to predict landbird images with water backgrounds in the query set. Optimizing a classifier's performance on these spuriousness-aware tasks facilitates the classifier to learn to be invariant to spurious correlations.

\subsection{SPUME Mitigates Reliance on Spurious Correlations}
\begin{figure}[t]
    \centering
    \includegraphics[width=\linewidth]{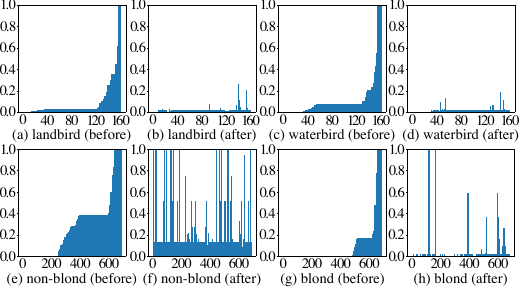}
    \caption{Spuriousness scores for all the class-attribute correlations before and after applying SPUME-BLIP to a classifier. The horizontal axes represent the indexes of detected attributes or class-attribute correlations, and the vertical axes represent the spuriousness scores. (a)-(d) Spuriousness scores on the Waterbirds dataset with landbird and waterbird classes. (e)-(h) Spuriousness scores on the CelebA dataset with non-blond and blond classes.}
    \label{fig:spuriousness-score}
\end{figure}

We calculated the spuriousness scores for all the detected class-attribute correlations before and after applying SPUME-BLIP to a classifier with the ResNet-50 backbone initialized with ImageNet pre-trained weights. We sorted the scores in the ``before" scenarios and kept the order in the corresponding ``after" scenarios. From Fig. \ref{fig:spuriousness-score}(a), (c), (e), and (g), we observe that the initial classifiers exhibit high reliance on the detected class-attribute correlations which have high spuriousness scores. After applying SPUME-BLIP to the classifiers on the Waterbirds dataset, we observe from Fig. \ref{fig:spuriousness-score}(b) and (d) that  the reliance on most of class-attribute correlations are mitigated and these correlations all have low spuriousness scores. On the CelebA dataset, which has more class-attribute correlations than the Waterbirds dataset, it becomes more challenging to mitigate the reliance on all these correlations. As observed from Fig. \ref{fig:spuriousness-score} (f) and (h), some correlations, which have low spuriousness scores initially, become highly spurious. Nevertheless, SPUME-BLIP can still mitigate the reliance on most of the class-attribute correlations having high spuriousness scores.
Moreover, since spuriousness scores are not directly incorporated into our optimization objective in \eqref{eq:objective}, the decrease in spuriousness scores demonstrates the effectiveness of our spuriousness-aware meta-learning strategy in mitigating the reliance on spurious correlations.

\subsection{Quantitative Evaluation}

\begin{table}[t]
\begin{tabular}{lcc}
\toprule
Method                                          & Worst-group acc. ($\uparrow$) & Acc. gap ($\downarrow$)    \\ \midrule
ERM                                             & 66.4            & 23.8        \\
LfF \cite{nam2020learning}     & 44.1            & 47.1        \\
CVaR DRO \cite{levy2020large}  & 62.0            & 33.2        \\
JTT \cite{liu2021just}         & 62.5            & 30.8        \\
DFR \cite{kirichenko2022last}  & 77.4            & 14.7        \\
DivDis \cite{lee2022diversify} & 81.0            & 9.7         \\ \midrule
SPUME-ViT-GPT2           & \textbf{85.9}$\pm$\textbf{0.2 }   & 6.9$\pm$0.8 \\
SPUME-BLIP                & 85.7$\pm$0.2  & \textbf{6.1}$\pm$\textbf{0.4} \\ \bottomrule
\end{tabular}%
\caption{Comparison of worst-group accuracy (\%) and accuracy gap (\%) on the Waterbirds dataset. All methods do not have access to ground-truth group labels.}\label{tab:waterbirds}
\end{table}

\begin{table}[t]
\begin{tabular}{lcc}
\toprule
Method                                          & Worst-group acc. ($\uparrow$) & Acc. gap ($\downarrow$)    \\ \midrule
ERM                                             & 45.7            & 49.8        \\
LfF \cite{nam2020learning}     & 24.4           & 60.7        \\
CVaR DRO \cite{levy2020large}  & 36.1           & 46.4        \\
JTT \cite{liu2021just}         & 40.6           & 47.4       \\
DFR \cite{kirichenko2022last}  & 46.0           & 49.8        \\
DivDis \cite{lee2022diversify} & 55.0           & 35.8         \\
MaskTune \cite{asgari2022masktune} &  78.0          &   13.3          \\\midrule
SPUME-ViT-GPT2            & 84.4$\pm$1.2  & 5.9$\pm$0.7 \\
SPUME-BLIP            & \textbf{86.0}$\pm$\textbf{1.0}  & \textbf{4.1}$\pm$\textbf{1.0} \\ \bottomrule
\end{tabular}%
\caption{Comparison of worst-group accuracy (\%) and accuracy gap (\%) on the CelebA dataset. All methods do not have access to ground-truth group labels.}\label{tab:celeba}
\end{table}

We compared our methods with prior methods on mitigating spurious correlations on the five datasets. On each of the datasets, we show the reported results of these methods when they are available and give the details of these methods in Appendix. 

For experiments on the Waterbirds and CelebA datasets, we aimed to simulate a more realistic learning scenario and thus did not provide group labels during model training, even though the two datasets provide group labels.  During testing, we used the group labels to formulate the \textit{worst-group accuracy} and calculated the \textit{accuracy gap} as the standard average accuracy minus the worst-group accuracy. The two metrics measure a classifier's robustness to \textit{specific} spurious correlations specified by the group labels, and our goal is to train the classifier to be robust to these spurious correlations without knowing them.  

Our methods, SPUME-ViT-GPT2 and SPUME-BLIP achieve the best worst-group accuracy and the best accuracy gap on the Waterbirds and CelebA datasets (Tables \ref{tab:waterbirds} and \ref{tab:celeba}), suggesting that our trained classifiers have strong and balanced prediction capability across different data groups. Note that the spurious attribute detection process proposed in Section \ref{sec:spurious-detection} could introduce biases present in VLMs into the detected spurious attributes. More specifically, biases in  different VLMs result in different sets of attributes. Consequently, SPUME simulates different sets of spurious correlations during meta-training. However,  this wouldn't be a significant concern.
Since our spurious attribute detection process can detect many distinctive attributes with well-established VLMs, SPUME can mitigate many potential spurious correlations. Thus, biases in VLMs won't significantly affect the effectiveness of our framework. 
We demonstrate this by showing that SPUME with two well-established VLMs are effective and have comparable performance across different datasets (Tables \ref{tab:waterbirds} and \ref{tab:celeba}). 
Moreover, SPUME-BLIP performs much better than SPUME-ViT-GPT2 on the CelebA dataset where BLIP detects approximately twice as many attributes as ViT-GPT2 does (Table \ref{tab:statistics-detected-spurious-attributes}), suggesting that detecting more attributes benefits  SPUME in training more robust classifiers.

\begin{table}[t]
\begin{tabular}{lc}
\toprule
Method                                          & Accuracy ($\uparrow$)    \\ \midrule
ERM      &      75.9                \\
REx \cite{krueger2021out}      &   74.3                  \\
Group DRO  \cite{sagawa2019distributionally}    &   77.6                  \\
JiGen \cite{carlucci2019domain}    &   85.0                 \\
Mixup \cite{zhang2017mixup}   &  80.3                   \\
CNBB  \cite{he2021towards}   &    78.2                  \\
DecAug \cite{bai2021decaug}  &   85.2               \\ 
SIFER \cite{tiwari2023overcoming} & 86.2$\pm$0.9 \\
\midrule
SPUME-ViT-GPT2            &  88.2$\pm$1.1  \\
SPUME-BLIP             &  \textbf{89.2$\pm$0.4} \\ \bottomrule
\end{tabular}%
\caption{Comparison of average accuracy (\%) on the NICO dataset. Most of the methods (DecAug, DRO, etc) use group information for training, while we do not use it.}\label{tab:nico}
\end{table}

The NICO dataset provides object-context correlations and aims to evaluate the out-of-distribution generalization capability of a classifier by testing it in new contexts. We did not use the provided correlations during training and calculated the standard average accuracy on the test set with new object-context correlations. SPUME-ViT-GPT2 and SPUME-BLIP outperform previous methods with higher average accuracies (Table \ref{tab:nico}). 

For the experiments on the ImageNet-9 which does not provide information on spurious correlations, we trained and tested our methods on the ImageNet-9 dataset. We also tested our methods on the ImageNet-A dataset which contains images representing various failure prediction modes in an ImageNet pre-trained classifier. The accuracy gap is calculated as the average validation accuracy on the ImageNet-9 dataset minus the average accuracy on the ImageNet-A dataset. Our methods achieve the best on ImageNet-A while well balancing between different prediction modes with the lowest accuracy gaps (Table \ref{tab:imagenet-9}). 

\begin{table}[t]
\resizebox{\linewidth}{!}{%
\begin{tabular}{lccc}
\toprule
Method                          & ImageNet-9  ($\uparrow$)  & ImageNet-A  ($\uparrow$) & Acc. gap ($\downarrow$) \\ \midrule
ERM                             & 90.8$\pm$0.6  & 24.9$\pm$1.1 & 65.9     \\
ReBias \cite{bahng2020learning} & 91.9$\pm$1.7  & 29.6$\pm$1.6 & 62.3     \\
LfF \cite{nam2020learning}      & 86.0          & 24.6         & 61.4     \\
CaaM \cite{wang2021causal}      & 95.7          & 32.8         & 62.9     \\
SSL+ERM \cite{kim2022learning}  & 94.2$\pm$0.1  & 34.2$\pm$0.5 & 60       \\
LWBC\cite{kim2022learning}      & 94.0$\pm$0.2  & 36.0$\pm$0.5 & 58       \\
SIFER \cite{pmlrv202tiwari23a}  & \textbf{97.8}$\boldsymbol{\pm}$\textbf{0.1} & 40.0$\pm$0.8 & 57.8     \\ \midrule
SPUME-ViT-GPT2           & 95.3$\pm$0.5          & \textbf{44.3}$\boldsymbol{\pm}$\textbf{0.8}         & \textbf{51.0}$\boldsymbol{\pm}$\textbf{1.1}     \\
SPUME-BLIP             & 95.5$\pm$0.2          & 42.5$\pm$0.8         & 53.0$\pm$0.7     \\ \bottomrule
\end{tabular}%
}
\caption{Comparison of average accuracy (\%) and accuracy gap (\%) on the ImageNet-9 and ImageNet-A datasets.}\label{tab:imagenet-9}
\end{table}

\subsection{Ablation Study}
\paragraph{Spuriousness-Aware Task Construction.} To evaluate the effectiveness of using VLMs to guide the construction of meta-learning tasks, we compared SPUME with SPUME-Random which uses randomly constructed tasks during training. We also included the classical ERM model and the ERM-Cosine model that uses cosine distance for predictions to compare with the meta-learning based approaches. We observe from Table \ref{tab:ablation-task} that switching to the cosine-distance-based classifier increases the robustness to spurious correlations. Moreover,
SPUME-Random outperforms ERM by 12.3\% in the worst-group accuracy and improves the accuracy gap by 13.3\%, demonstrating that meta-learning is a promising approach to improve the robustness to spurious correlations. Additionally, using spuriousness-aware meta-learning tasks constructed with the VLMs (BLIP and ViT-GPT2) can further improve robustness to spurious correlations. Specifically, SPUME-BLIP achieves 7.0\% and 4.4\%  increments over SPUME-Random in the worst-group accuracy and accuracy gap, respectively, and SPUME-ViT-GPT2 achieves 7.2\% and 3.6\% increments in the two metrics.

\begin{table}[t]
\begin{tabular}{ccc}
\toprule
Method          & Worst-group acc ($\uparrow$) & Acc. gap ($\downarrow$) \\ \midrule
ERM             & 66.4                         & 23.8                    \\
ERM-Cosine & 75.5  & 17.5\\ \midrule
SPUME-Random     & 78.7$\pm 0.9$                & 10.5$\pm 0.8$           \\ 
SPUME-BLIP     & 85.7$\pm 0.2$                & \textbf{6.1}$\boldsymbol{\pm}$\textbf{0.4}            \\
SPUME-ViT-GPT2 & \textbf{85.9}$\boldsymbol{\pm}$\textbf{0.3}                 & 6.9$\pm$0.8             \\ \bottomrule
\end{tabular}%
\caption{Worst-group accuracy and accuracy gap comparisons between meta-learning based methods with spuriousness-aware (SPUME-BLIP and SPUME-ViT-GPT2) and random (SPUME-Random) task constructions, and ERM-trained models on the Waterbirds dataset.}\label{tab:ablation-task}
\end{table}

\paragraph{Different Designs of the Spuriousness Metric.} 
We have given our design of spuriousness metric in Eq. \eqref{eq:spuriousness-score}. Here, we explore other possible design choices shown in Table \ref{tab:ablation-spuriousness-metrics}, where $\delta=J(\mathcal{D}_{tr}^{( y,a)};f_{\theta})-J(\mathcal{D}_{tr}^{( y,\hat{a})};f_{\theta})$, $\eta=J(\mathcal{D}_{tr}^{( y,a)};f_{\theta})/J(\mathcal{D}_{tr}^{( y,\hat{a})};f_{\theta})$,  $J(\cdot;\cdot)$ is the accuracy measure used in Eq. \eqref{eq:spuriousness-score}, and ``Constant" represents that we assign the same score for all the detected attributes. Our method SPUME-BLIP works well with different spuriousness metrics and still outperforms the baselines we compared in Table \ref{tab:waterbirds}. Moreover, our method works well with non-negative spuriousness metrics as SPUME with tanh(abs(log($\eta$))) or abs($\delta$) performs better than with the other two metrics.
\begin{table}[t]
\begin{tabular}{lcc}
\toprule
Metric               & Worst-group acc. ($\uparrow$) & Acc. gap ($\downarrow$) \\ \midrule
tanh(abs(log($\eta$))) &  \textbf{85.7}$\pm$\textbf{0.2}    &    \textbf{6.1}$\pm$\textbf{0.4}      \\
abs($\delta$)      &    85.5$\pm$0.2  &  6.3$\pm$0.3   \\
Constant     &    85.1$\pm$0.2  &  6.7$\pm$0.3   \\
  tanh(log($\eta$))          &    84.8$\pm$0.2   &  7.3$\pm$0.4       \\
$\delta$                &    84.5$\pm$0.5             &   7.4$\pm$0.9       \\ \bottomrule
\end{tabular}%
\caption{Analysis on different designs of spuriousness metrics. We tested SPUME-BLIP on the Waterbirds dataset.\label{tab:ablation-spuriousness-metrics}}
\end{table}

\begin{figure}
    \centering
    \includegraphics[width=0.85\linewidth]{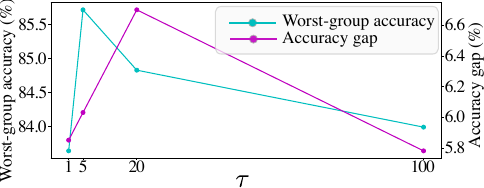}
    \caption{Worst-group accuracy and accuracy gap comparisons between SPUME-BLIP with different $\tau$'s on  Waterbirds.}
    \label{fig:waterbirds-blip-temp-ablation}
\end{figure}

\paragraph{Scaling Parameter of the Centroid-Based Classifier.} 
We analyzed how the scaling parameter $\tau$ of the centroid-based classifier (Eq. \eqref{eq:centroid-classifier}) affects the performance of SPUME. Fig. \ref{fig:waterbirds-blip-temp-ablation} shows the worst-group accuracies and accuracy gaps of SPUME-BLIP with different $\tau$'s on the Waterbirds dataset. A very large or small $\tau$, e.g., $\tau=100$ or $\tau=1$, harms to robustness of the trained classifiers. In practice, we set $\tau$ to be in the range from 5 to 50.

\paragraph{Effects of Using VLMs.} Although SPUME uses VLMs for data preprocessing, the robustness does not directly come from the outputs of VLMs. To show this, we added an additional layer after the backbone to predict detected attributes for each image, acting as a regularization. We then fine-tuned the whole model on the Waterbirds and CelebA datasets, respectively. The worst-group accuracies on the two datasets are 71.7\% and 47.2\%, respectively, which are close to ERM trained models. Therefore, the attributes themselves do not provide useful regularization on the robustness of the classifier.  Moreover, directly using VLMs for predictions requires a completely different inference pipeline and is not as effective as our proposed SPUME. Details are provided in Appendix.

\section{Conclusion}
We proposed a novel framework to train a classifier to be robust against spurious correlations in settings where spurious correlations are not known or specified. We first adopted a pre-trained VLM to automatically extract text-format attributes from a target dataset. Then, we quantified the spuriousness of the correlations between detected attributes and class labels using a spuriousness metric. To effectively mitigate multiple detected spurious correlations, we adopted a meta-learning strategy which iteratively meta-trains a classifier on multiple meta-learning tasks constructed to represent various class-attribute correlations with high spuriousness values. Our framework, SPUME, mitigates many highly spurious correlations in training samples and performs the best under different robustness measures on five benchmark datasets. In the future, we aim to explore more capable VLMs and combine other approaches, e.g., customized data augmentations, for mitigating a model's reliance on a wider range of spurious correlations.

\begin{acks}
This work is supported in part by the US National Science Foundation under grants 2217071, 2213700, 2106913, 2008208, 1955151.
\end{acks}

\bibliographystyle{ACM-Reference-Format}
\balance
\bibliography{sample-base}
\appendix

\section{Appendix}
\subsection{Datasets}
Table \ref{tab:dataset-statistics} depicts detailed statistics for all datasets.  For Waterbirds and CelebA datasets, we give the number of training, validation, and test images in each group specified by classes and attributes. For example, the group label (landbird, land) in the Waterbirds dataset has 3498 training images which are all landbird and have land backgrounds. NICO provides context labels as spurious attributes. ImageNet-9 and ImageNet-A datasets do not have clear group partitions specified by the class and attribute associations.

\begin{table}[H]
\resizebox{\linewidth}{!}{%
\begin{tabular}{lclccc}
\toprule
\multicolumn{1}{l}{\multirow{2}{*}{Dataset}} & \multicolumn{1}{l}{\multirow{2}{*}{\begin{tabular}[c]{@{}c@{}}Number of \\ classes\end{tabular}}} & \multicolumn{1}{c}{\multirow{2}{*}{$\langle$class, attribute$\rangle$}} & \multicolumn{3}{c}{Number of images} \\ \cmidrule{4-6} 
\multicolumn{1}{l}{}                         & \multicolumn{1}{l}{}                                   & \multicolumn{1}{l}{}                                                  & Train       & Val        & Test      \\ \midrule
\multirow{4}{*}{Waterbirds}                  & \multirow{4}{*}{2}                                     & $\langle$landbird, land$\rangle$                                        & 3,498       & 467        & 2,255     \\
                                             &                                                        & $\langle$landbird, water$\rangle$                                       & 184         & 466        & 2,255     \\
                                             &                                                        & $\langle$waterbird, land$\rangle$                                       & 56          & 133        & 642       \\
                                             &                                                        & $\langle$waterbird, water$\rangle$                                      & 1,057       & 133        & 642       \\ \midrule
\multirow{4}{*}{CelebA}                      & \multirow{4}{*}{2}                                     & $\langle$non-blond, female$\rangle$                                     & 71,629      & 8,535      & 9,767     \\
                                             &                                                        & $\langle$non-blond, male$\rangle$                                       & 66,874      & 8,276      & 7,535     \\
                                             &                                                        & $\langle$blond, female$\rangle$                                         & 22,880      & 2,874      & 2,480     \\
                                             &                                                        & $\langle$blond, male$\rangle$                                           & 1,387       & 182        & 180       \\ \midrule
NICO                                         & 10                                                     & $\langle$object, context  $\rangle$                                                                   & 10298        & 642       & 894      \\ \midrule
ImageNet-9                                   & 9                                                      & -                                                                     & 54,600      & 2,100      & -         \\ \midrule
ImageNet-A                                   & 9                                                      & -                                                                     & -           & -          & 1087      \\ \bottomrule
\end{tabular}%
}\caption{Detailed statistics of the 5 datasets. $\langle$class, attribute$\rangle$ represents a spurious correlation between a class and a spurious attribute. ``-" denotes not applicable.}\label{tab:dataset-statistics}
\end{table}

\begin{table}[H]
\small
\begin{tabular}{ccc}
\toprule
\multirow{2}{*}{Class}    & \multicolumn{2}{c}{Contexts}  \\ \cmidrule{2-3}
          &   Validation   & Test\\ \midrule
dog      & running & in\_street        \\
cat      & on\_tree & in\_street     \\
bear     & on\_tree & white          \\
bird     & on\_shoulder & in\_hand  \\
cow      & spotted & standing    \\
elephant & in\_circus & in street      \\
horse    & running & in\_street \\
monkey   & climbing & sitting       \\
rat      & running & in\_hole         \\
sheep    & at\_sunset & on\_road    \\ \bottomrule
\end{tabular}%
\caption{Classes and their associated contexts in the NICO datasets. Contexts not shown in the table are used in the training set.}\label{tab:class-context-nico}
\end{table}
NICO \cite{he2021towards} is a real-world dataset
for out-of-distribution robustness. We used its Animal subset containing 10 object classes and 33 context labels. Following the setting in \cite{bai2021decaug,tiwari2023overcoming}, we  split the dataset into training, validation, and test sets with each set having unique contexts.   Table \ref{tab:class-context-nico} gives the allocation of the contexts for the 10 classes.

The ImageNet-9 dataset \cite{bahng2020learning} is a subset of ImageNet. It has 9 super-classes, i.e., Dog, Cat, Frog, Turtle, Bird, Primate, Fish, Crab, Insect, which are obtained by merging similar classes from ImageNet. ImageNet-A contains real-world images
that are challenging to the image classifiers trained on standard ImageNet.  We extract images of the 9 super-classes from the ImageNet-A dataset and use these images as the test data.

\subsection{Experimental Details}
\paragraph{VLM Settings.} For both ViT-GPT2 and BLIP, we set the maximum length of the sequence to be generated as 16 and the number of beams for beam search to 4.

\paragraph{Training Details.} We initialize ResNet-50 and ResNet-18 using ImageNet pre-trained weights. Standard data augmentations, i.e., \texttt{RandomResizedCrop} and \texttt{RandomHorizontalFlip} are used during model training. We use an SDG optimizer with a momentum of 0.9 and a weight decay of $10^{-4}$ during meta-training. The detailed training configurations are shown in Table \ref{tab:training-details}.
\begin{table*}[t]
\centering
\begin{tabular}{ccccccc}
\toprule
Dataset    & Learning rate &\begin{tabular}[c]{@{}c@{}}Learning rate \\ scheduler\end{tabular} & \begin{tabular}[c]{@{}c@{}}Number of tasks \\ per epoch\end{tabular} & Training epochs & $\tau$ & \begin{tabular}[c]{@{}c@{}}Model selection\\ metric\end{tabular}               \\ \midrule
Waterbirds & 1e-3          & Cosine Annealing        & 80                                                                     & 100     & 5 & $Acc_{pu}$ \\
CelebA     & 1e-3          & Cosine Annealing         & 80                                                                     & 100     & 5 & $Acc_{pu}$ \\
NICO       & 5e-3          & Cosine Annealing         & 80                                                                     & 50     & 10 & Validation accuracy                                                           \\
ImageNet-9 & 1e-3          & Cosine Annealing         & 80                                                                     & 50     & 50 & Validation accuracy                                                            \\ \bottomrule
\end{tabular}%
\caption{Hyperparameter settings and model selection criteria for SPUME training on the Waterbirds, CelebA, NICO, and ImageNet-9 datasets. $Acc_{pu}$ denotes pseudo unbiased validation accuracy.}\label{tab:training-details}
\end{table*}

\subsection{Baselines}\label{sec:baselines}
We briefly summarize and describe the baselines which are compared in the experiments:

\noindent \textbf{Group DRO \cite{sagawa2019distributionally}} proposes to train the models on the worst-case loss over a set of predefined groups.

\noindent \textbf{ReBias \cite{bahng2020learning}} proposes a novel framework to train a de-biased representation by encouraging it to be different from a set of biased representations.

\noindent \textbf{REx \cite{krueger2021out}} proposes a min-max algorithm to optimize for the worst linear combination of risks on different environments.

\noindent \textbf{LfF \cite{nam2020learning}} proposes a failure-based debiasing scheme by training a pair of neural networks: the first network to be biased by repeatedly amplifying its ``prejudice" and debias the training of the second network by focusing on samples that counter the first network.

\noindent \textbf{CVaR DRO \cite{levy2020large}} is an algorithm for distributionally robust optimization of convex losses with conditional value at risk (CVaR) and $\chi^2$ divergence uncertainty sets. 

\noindent \textbf{JTT \cite{liu2021just}} proposes a simple two-stage approach that first trains a standard ERM model and then trains a second model by upweighting the training examples misclassified by the first model.

\noindent \textbf{DFR \cite{kirichenko2022last}} retrains the last linear layer on a small held-out dataset with balanced groups of data.

\noindent \textbf{CaaM \cite{wang2021causal}} learns causal features that are robust in any confounding context and self-annotates the confounders in an unsupervised fashion.

\noindent \textbf{LWBC / SSL+ERM \cite{kim2022learning}} employs a committee of classifiers as an auxiliary module that identifies bias-conflicting data and assigns large weights to them when training the main classifier. SSL+ERM is another approach proposed in this paper that uses self-supervised representation as the frozen backbone of the committee and the main classifier.

\noindent \textbf{MaskTune \cite{asgari2022masktune}} employs an interpretation-based masking strategy that mitigates over-reliance on spurious features. It forces the trained model to explore new features during a single epoch fine-tuning by masking previously discovered features.

\noindent \textbf{DivDis \cite{lee2022diversify}} is a simple two-stage framework for identifying and resolving ambiguity in data. It  first learns a diverse set of hypotheses and then disambiguates them by selecting one of the discovered functions using additional information (e.g. target labels).

\noindent \textbf{JiGen \cite{carlucci2019domain}} jointly classifies objects and solves unsupervised jigsaw tasks.

\noindent \textbf{Mixup \cite{zhang2017mixup}} trains a neural network on convex combinations of pairs of examples and their labels to alleviate memorization and sensitivity to adversarial examples in deep neural networks.

\noindent \textbf{CNBB \cite{he2021towards}} is a non-independent and identically distributed (Non-I.I.D) learning method that is based on batch balancing inspired by causal inference.

\noindent \textbf{DecAug \cite{bai2021decaug}} proposes a semantic augmentation and feature decomposition approach to disentangle context
features from category-related features.

\noindent \textbf{SIFER \cite{pmlrv202tiwari23a}} automatically identifies and suppresses easily-computable spurious features in lower layers of the network and allows the higher layers of the network to extract and utilize more meaningful representations.


\subsection{Analyzing the Effects of Using VLMs}
\paragraph{Using the Outputs of VLMs as Regularization.}  We added a linear layer with weights $\mathbf{W}_A\in\mathbb{R}^{|\mathcal{A}|\times D}$ and bias $\mathbf{b}_A\in\mathbb{R}^{|\mathcal{A}|}$ after the backbone to predict the detected attributes for each image, i.e.,
\begin{align}\label{eq:erm-objective-reg}
    \tilde{\theta} &=\arg \min_{\theta}\mathbb{E}_{(x,y)\in \mathcal{D}_{\text{tr}}}\ell(f_{\theta}(x),y)+\sum_{a\in\psi(\phi(x))}\ell'(f'_{\theta'}(x),a)
\end{align}
where $f'_{\theta'}(x)=\mathbf{W}_{A}h_{\theta_1}(x)+\mathbf{b}_A$, and $\ell'(\cdot,\cdot)$ is a binary entropy loss. We trained the whole model on the Waterbirds and CelebA datasets, respectively.  If the attributes contain information effective in improving a classifier's robustness to spurious correlations, we will observe improved performance after training. However, the worst-group accuracies on the Waterbirds and CelebA datasets are 71.7\% and 47.2\%, respectively, which are only slightly better than those of ERM and fall far behind the results of SPUME. Therefore, the detected attributes from the VLM alone do not contain information effective for improving a classifier's robustness to spurious correlations.

\paragraph{Directly Using VLMs for Predictions.} Although the goal of this paper is to learn a classic and resource-light classifier that is robust to spurious correlations, we explored the scenario when BLIP is directly used for prediction with modifications on the inference paradigm. Specifically, we used text embeddings of the sentences with the template ``a photo of \texttt{class\_label}" (``a person with \texttt{hair\_color} hair" for CelebA) from BLIP as the classifier weights and calculated the cosine similarity between an image embedding and these weights in the shared embedding space of BLIP. We predicted the label such that its corresponding sentence has the highest similarity to the image embedding. The worst group accuracies on the Waterbirds and CelebA datasets are 1.17\% and 29.71\% respectively. The average accuracies on the NICO, ImageNet-9, and ImageNet-A datasets are  14.30\%, 13.43\%, and 9.20\%, respectively. Directly using the VLM without carefully tuning the inference pipeline performs much worse than our proposed method. In contrast, our proposed method SPUME exploits the attributes provided by VLMs in a novel way for significant improvement in the robustness of a classifier to spurious correlations.

\end{document}